# A Comprehensive Survey on Deep-Learning-based Vehicle Re-Identification: Models, Data Sets and Challenges

Ali Amiri, Aydin Kaya, and Ali Seydi Keçeli

*Abstract*—**Vehicle re-identification (ReID) endeavors to associate vehicle images collected from a distributed network of cameras spanning diverse traffic environments. This task assumes paramount importance within the spectrum of vehicle-centric technologies, playing a pivotal role in deploying Intelligent Transportation Systems (ITS) and advancing smart city initiatives. Rapid advancements in deep learning have significantly propelled the evolution of vehicle ReID technologies in recent years. Consequently, undertaking a comprehensive survey of methodologies centered on deep learning for vehicle re-identification has become imperative and inescapable. This paper extensively explores deep learning techniques applied to vehicle ReID. It outlines the categorization of these methods, encompassing supervised and unsupervised approaches, delves into existing research within these categories, introduces datasets and evaluation criteria, and delineates forthcoming challenges and potential research directions. This comprehensive assessment examines the landscape of deep learning in vehicle ReID and establishes a foundation and starting point for future works. It aims to serve as a complete reference by highlighting challenges and emerging trends, fostering advancements and applications in vehicle ReID utilizing deep learning models.**

## I. INTRODUCTION

VEHICLES are the most popular and important part of social life. Recent advances in vehicle-related technologies, such as vehicle detection, vehicle type recognition, vehicle tracking, vehicle retrieval, etc., have prompted essential roles in realising Intelligent Transportation Systems (ITS) and smart cities [1-5]. Within this spectrum of technologies, searching for a specific vehicle trajectory and exploring its movements holds heightened significance in bolstering public safety within smart city frameworks [6]. As Illustrated in Fig. 1, this endeavor involves extracting global and local features from vehicle images, coupled with pertinent auxiliary attributes, including color, type, brand, and spatiotemporal data. These extracted features are then used to compare gallery images, retrieve similar images with query images, narrow the scope of retrieval, and ultimately augment the quality of outcomes. The key element of this process involves extracting and comparing of features extracted from vehicle images, commonly known as vehicle ReID [6-8]. Vehicle ReID is designed to identify a specific vehicle within an extensive repository of vehicle images captured by various cameras at disparate moments. It is particularly pronounced in expansive ITS and finds application across various video surveillance scenarios, including locating lost vehicles, conducting cross-regional tracking of specific vehicles, and more [9, 10].

Recently, with the rapid advances in deep learning techniques and its success in high-performance automatic object detection [11], vehicle ReID has attracted the attention of many researchers and industrialists. According to the review investigated over several research papers, our taxonomy of the deep learning-based vehicle ReID methods is illustrated in Fig. 2. As can be seen, two key categories under vehicle Re-ID methods are supervised approaches and unsupervised approaches.

In supervised methods, some researchers have mainly concentrated on learning distinctive visual features from vehicle images as a classification problem, and others have focused on deep metric learning by a loss function.

To learn of visual features, generally, Convolutional Neural Network (CNN) and its variations have been widely adopted to identify global features from vehicle images [12]. Also, some methods, such as [13] and [14], have been suggested to combine the feature maps of different layers against the final output of CNN. These methods only focus on the global features of the vehicle without taking into account its abundant details and thus are not suitable for discriminating visually similar vehicles. To pursue this, some work has been done to integrate local and global features. Local features such as vehicle brands or decorations from different views can reflect additional details. These methods principally emphasize learning local features from image partitions [15-21], considering relations between part regions [22-25], or learning interaction between local and global features [26]. Lately, the capabilities of transformers have been adopted to learn global and local features from vehicle images [26-29]. In addition to vehicle images, some methods have employed knowledge-based information including but not limited to vehicle attributes [30, 31] such as

Ali Amiri is with the Computer Engineering Department, Hacettepe University, Ankara, Turkey, on leave from the Computer Engineering Group, Faculty of Engineering, University of Zanjan, Zanjan, CO 45371-38791 Iran (e-mail: a_amiri@znu.ac.ir).

Aydin Kaya and Ali Seydi Keçeli are with the Computer Engineering Department, Hacettepe University, Ankara, Turkey (e-mail: aydinkaya@cs.hacettepe.edu.tr and aliseydi@cs.hacettepe.edu.tr).

vehicle color, type, brand, and spatiotemporal features such as vehicle trajectories [32].

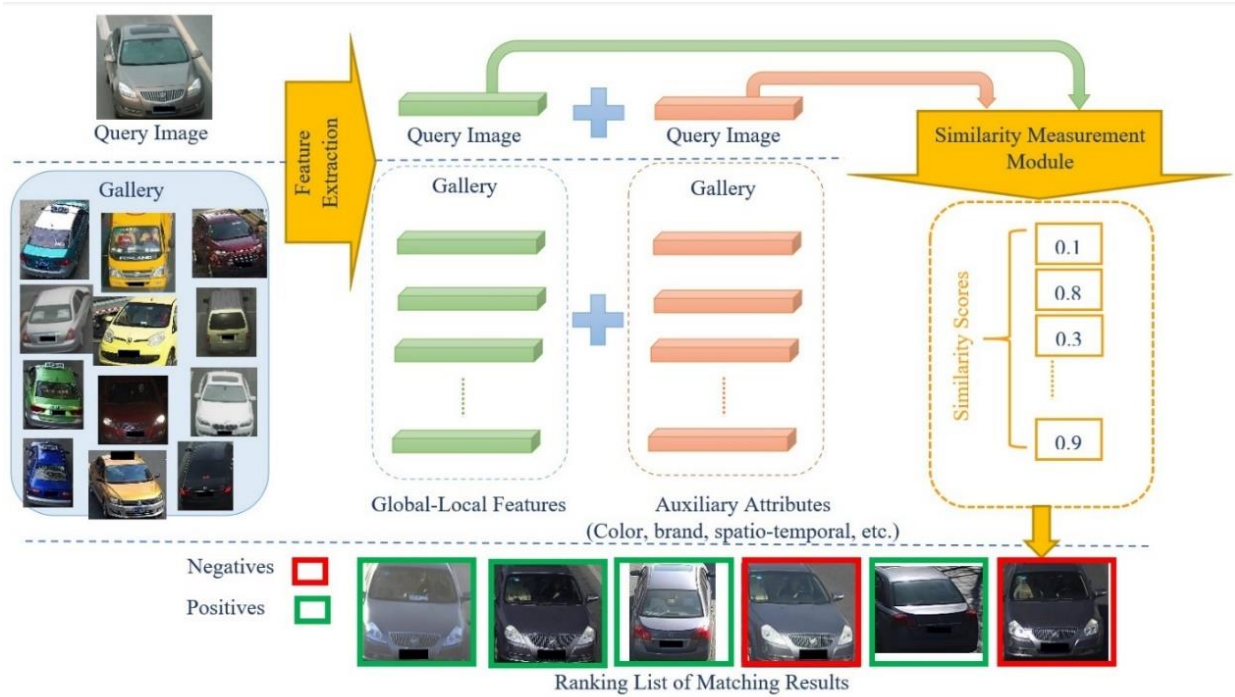

**Fig. 1.** Vehicle ReID description. The vehicle image's global and local features and additional auxiliary attributes are extracted and used to compare with gallery images to retrieve the ranking output list.

Deep metric learning aims to obtain a multidimensional feature space through deep models so that instances with identical class labels are close and instances of dissimilar class labels are far from each other. The two significant loss functions employed for metric deep learning are contrastive loss and triplet loss. Regarding the recent studies, triplet loss is superior to contrastive loss in vehicle ReID tasks [33]. Furthermore, certain adaptations of these loss functions have been specifically tailored to accommodate the constraints associated with vehicle ReID [8, 34].

Unsupervised approaches attempt to completely discover the appropriate information from the data without any class label, which can be separated into two groups: unsupervised domain adaptation and fully unsupervised. The former methods concentrate on applying some modifications of the adversarial network (GAN), such as PTGAN [35], SPGAN [36], and CycleGAN [37], to generate artificial images from the source domain with identical class labels. These images are consumed in a supervised manner to train deep networks. The latter mainly aims to develop clustering algorithms and training policies to discover appropriate information only from the target domain data without considering other supplementary information. Unlike supervised approaches, these methods infer vehicle ReID from unlabeled data without annotation, making it more suitable and enforceable for real-world scenarios [38-40].

While most recent achievements have focused on improving the performance of vehicle ReID models and solving its issues, less research has focused on comprehensively reviewing these developments and improvements. To the best of our knowledge, no thorough review of a comprehensive study has been presented to encapsulate and discover all aspects of this topic, excluding some surveys on supervised methods, such as [41], [42] and [43]. Therefore, it is necessary to review the current state of the art to supply counsel for future research on this subject.

In this review, we present a comprehensive survey of vehicle ReID using deep learning methods, introduce the general categorization of these methods, including supervised and unsupervised methods, provide a review of the existing research of these categories, explain the well-known data sets and evaluation criteria, and describe the challenges and possible directions of future researches. This review examines the current state of deep learning-based vehicle ReID. It serves as a foundation and a starting point for future work, pointing out the corresponding challenges and tendencies. The final result will be useful for future developments and applications of vehicle ReID with deep learning models.

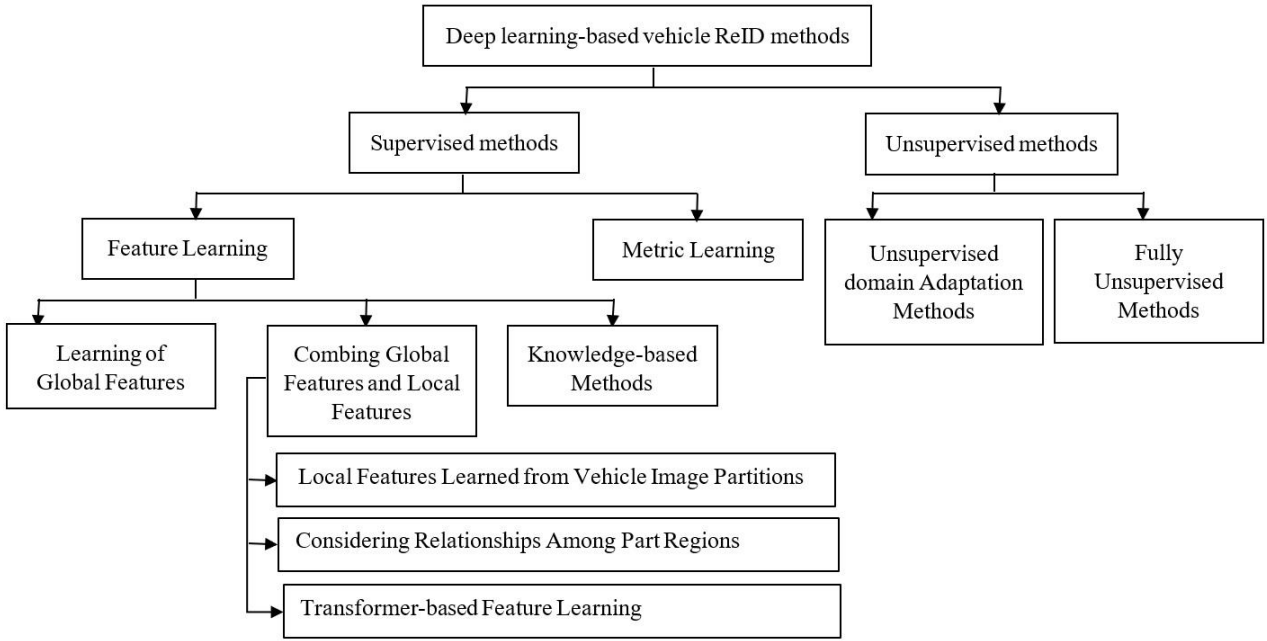

**Fig. 2.** A hierarchical taxonomy of the deep learning-based vehicle ReID methods.

The structure of the rest of this paper is as follows. In Section 2, we formulate the vehicle re-identification problem mathematically. Section 3 overviews existing supervised approaches to the vehicle ReID problem. Also, a review of related unsupervised approaches is explained in Section 4. An overview of well-known standard datasets and their comparison is discussed in Section 5. Evaluation strategies and performance measure criteria are studied in Section 6. Then, Section 7 talks about the challenges and upcoming research guidelines. Finally, the conclusion is presented in Section 8.

## II. Problem Formulation

Define $G = \{(I_k^{v_k}, L_k) | k = 1, 2, ..., N\}$ as an arbitrary large-scale gallery of vehicle images captured by different cameras with various directions in a surveillance system, where $I_k$, $L_k$ and $v_k$ are the $k^{th}$ vehicle image, its corresponding identity label, and its direction, respectively.

The vehicle orientation is an important item that impresses the efficiency of the vehicle ReID model because there are usually major changes in the appearance of the vehicle when it is captured from a different direction. The extracted appearance features must be able to utilise subtle details, such as the decorations behind the windshield and the air inlets of the engine, to differentiate between visually similar vehicles captured from the same angle. The vehicle ReID model needs to extract features that are both orientation-invariant and similar. This is essential for comparing images of the same vehicle with different orientations but the same identity label. As a result, we have considered the orientation (with the symbol $v_k$) as an independent parameter in our formulation. We define the identity label sameness criterion between two images as follows:

$$s_{i,j} = \begin{cases} 1 & \text{if } I_i \text{ and } I_j \text{ have same identity label} \\ 0 & \text{Otherwise} \end{cases} \quad (1)$$

In recent works, researchers aim to develop a complex deep-learning model to extract the visual global and local features from vehicle images to describe their visual characteristics. We formulate the deep learning-based appearance feature extraction module by the function $\Gamma$ as follows:

$$f_k = \Gamma\big(I_k^{v_k}, \theta_{v_k}\big), \; for \; k = 1, 2, ..., N, \quad (2)$$

where $\theta_k$ is the attribute vector of the $k^{th}$ vehicle from the viewpoint $v_k$, and $f_k$ is its normalized extracted appearance feature map.

From the recent efforts perspective, vehicle Re-ID still suffers from two critical challenges: 1) vehicle images with the same identity label, but different viewpoints or capturing time have dissimilar appearance features, which causes intra-instance dissimilarity issues. 2) vehicles with different identities but with the same attributes (color, type, brand, etc.), have similar appearance features and cause inter-instance similarity issues. To consider these challenges in the problem formulation, we can assume that the vehicle ReID problem seeks a deep learning method to extract suitable global and local features as $f_k$ in Eq. (1), and a loss metric function such that for every two arbitrary vehicle images $I_i^{v_i}$ and $I_j^{v_j}$ in the gallery, the distance between $f_i$ and $f_j$ should be minimized when $s_{i,j} = 1$ and maximized for $s_{i,j} = 0$. The objective function of this optimization problem is formulated using different loss functions. For example, by utilizing contrastive loss [50], the objective function will be as follows:

$$\mathcal{L} = \frac{1}{N^2} \Sigma_{i,j=1}^{N} \{s_{i,j} \|f_i - f_j\|_2^2 + (1 - s_{i,j}) \max(0, 1 - \|f_i - f_j\|_2^2)\}, \quad (3)$$

As soon as the deep learning model is effectively trained to extract the distinctive features of vehicles, the query image can be simply compared to the gallery to obtain a ranking. Therefore, the crucial component to effectively solve the vehicle ReID problem is designing and learning the function $\Gamma(.)$ through a suitable deep learning-based model. The coming sections give an overview of recent efforts to deal with this problem.

## III. SUPERVISED VEHICLE RE-IDENTIFICATION

As mentioned earlier, supervised approaches mainly focus on learning the discriminative appearance model as a visual feature from vehicle images using supervised learning techniques. This is achieved by two different policies: feature learning and metric learning. The former employs a deep learning model to treat the vehicle ReID problem by learning robust and discriminative features from vehicle images. In contrast, the latter focuses on designing distance and loss functions such that instances with identical class labels are close and instances of dissimilar class labels are far from each other. This section provides a comprehensive overview of these policies.

### A. Feature Learning

Recent research has mostly focused on adopting a variation of CNN to extract global or local features and combine them with an aggregation module to re-identify vehicles through feature learning. According to the aggregation module mechanism, these methods can be classified into three categories: learning global features, combining global and local features, and knowledge-based methods. In the following subsections, an overview of recent research on these categories and their advantages and disadvantages are discussed in detail.

#### 1) **Learning of Global Features**

Global feature learning methods typically habit a poling layer in CNN to compact global features as a big picture of the vehicle and disregard the local features. The method in [10] applied an end-to-end two-branch CNN framework called Deep Linked Distance Learning (DRDL) to map raw vehicle images into a Cartesian domain to allow comparison of vehicle similarity with $L_2 - norm$ distance measure. This framework simultaneously uses Triplet loss in Cartesian space as a cluster loss function to minimize the distance between vehicle images with the same identity labels and maximize the distance between images with different identities. In [44], the authors presented an extended triplet-wise training mechanism for vehicle ReID by joint learning of triplet loss and classification-oriented loss for vehicle ReID. A modified CNN is adopted, and a triplet sampling method is presented to generate triplet images as input to the network. The network is trained by joint learning to learn the visual features of the images. The network maps the input images into Cartesian space, and also similar to [10], the $L_2 - norm$ distance can be used to measure the similarity of vehicle images.

Some extensions of this idea have been presented to consider more aspects of the intrinsic challenges and improve the real-world generalization capacity of ReID models. In [13], a CCN network and a four-part coarse-to-fine ranking loss function have been adopted to consider all the significant challenges of the problem together and extract the appearance features of the vehicles. In the first step, the classification loss function was used to cluster vehicles with the same models and isolate ones with different models. Then, the coarse-grained ranking loss was introduced to improve the discrimination among vehicles of different models while maintaining some differences among vehicles of the same model. The discrimination among different vehicles of the same models was also formulated by fine-grained ranking loss. Also, pairwise loss was utilized to put the samples of the same vehicle as close as possible. Finally, they applied the stochastic gradient descent method to optimize the weights of the CNN network and train the visual appearance model. Although this model has formulated the main issues of the vehicle ReID, the model validation dataset is not suitable, and some challenges, including different views, occlusion, day and night, and different weather conditions, have not been assessed.

In addition to vehicle images, some works attempted to extract spatio-temporal attributes and associations among vehicle images in different views to improve the effectiveness of the vehicle ReID model. In [45], the relationship between vehicle images is formulated as multiple grains. Also, two methods, generalised pairwise and multi-grain-based list ranking, are proposed to improve the efficiency of vehicle retrieval problems. These methods have been implemented by CNN to extract global features.

The authors in [46] developed an integrated CNN-based framework to discover distinct visual representations for vehicle images. This framework effectively integrates four diverse subnetworks, including identification, attribute recognition, verification, and triplet, to learn various features and relationships between samples. The first two subnetworks extract detailed features of individual instances, while the next two focus on sample relationships. More precisely, verification and triplet limit the relationship between two samples and three samples, respectively. Finally, to train the framework, they proposed a method to optimize four objective functions of these subnetworks together.

In addition to the CNN-based appearance feature and auxiliary features such as color, type, brand, and texture, some researchers have focused on other specific features such as license plate and spatiotemporal information to make the retrieved results more accurate and enhance the re-ranking mechanism. The authors of [47] propose a multi-level deep network as a coarse filter to obtain vehicle visual appearance features. Then, they extend the coarse filter by adding attributes such as vehicle texture, color, and type. After that, license plate recognition was applied to make the search more precise. The spatiotemporal relationship between the query images and the gallery is adapted to modify the ranking of the retrieved images [48].

Recently, some research has been completed to consider the multi-resolution nature of vehicle images, which different cameras may capture. In [14], a two-stage deep model was presented to discover distinct visual features from multi-resolution images. In the first stage, a multi-branch network was developed to obtain particular properties of different scales. Each branch was composed of a similar structure CNN network to produce its scale-specific visual feature. These features provide an input to an integrated network to produce the final visual appearance model. The interaction between the output of the two stages has been utilized to raise the model's efficiency. In addition to the multi-resolution view, some efforts have been completed considering the attention mechanism and global appearance features to capture more informative key points. These works can be placed in the Combining Global and Local Features category. However, some of these methods do not divide the vehicle image into meaningful local parts and get only some global key points from vehicle images. We consider them in the current section. For example, in [49] a two-branch deep model is presented to extract global and local features. In the first branch, a global appearance model is extracted by a multi-level CNN-based network. Also, inspired by the works of [50] and [51], a fully two-stage CNN-based attention method is established to extract key points. These features are concatenated and post-processed to extract the final visual feature. Their evaluation results confirm the attention mechanism's effectiveness in overcoming the vehicle ReID issues.

In summary, despite the mentioned valuable efforts, global feature learning considers the overall view of vehicles and ignores the discriminative local features essential for vehicle ReID; as a result, these vehicle ReID models cannot achieve acceptable performance. Therefore, learning global features alone is not effective enough in vehicle Re-Id, and the local features should be considered to reflect vehicle details.

### 2) Combining Global and Local Features

Color, brand, type, and model similarities are pandemic attributes between vehicles, and as a result, vehicle ReID based on global visual features alone seems impossible, and local areas such as trim and inspection stickers affixed to the windshield may be more effective.

Additionally, due to the several challenges of camera angle changes, different weather and lighting conditions, and the similarity between vehicles, complicating deep models to learn global visual features without considering local features cannot effectively overcome the challenges of vehicle ReID [28]. This subsection provides a comprehensive review of the taxonomy of local feature learning methods and their most prominent models.

Figure (2) shows that local feature learning methods are divided into three categories. Some researchers have only focused on static or dynamic segmenting of vehicle images to extract local features. However, some works continue to investigate the relationship between segmented parts to improve the generalization capacity of the model. Finally, recent efforts have taken into account the interaction between global and local features.

- #### Local Features Learned from Vehicle Image Partitions

Local feature learning methods can be encapsulated into constant spatial partitioning and part detection methods. Besides convolutional neural networks, attention mechanisms and transformers play a significant role in learning local features in both methods.

Constant spatial partitioning methods usually split the feature maps into several pieces in the horizontal or vertical direction and then pool each piece separately. For example, in [19] a bifurcated deep model, including stripe-based and attribute-aware, was presented to simultaneously consider local and global features. The former consists of an average pooling layer combined with a dimensionality reduction convolutional layer to discover local visual features. At the same time, the latter extracts a global feature map by monitoring vehicle attribute labels to distinguish similar identities with different attribute annotations. Finally, the visual feature map of the vehicle image was constructed by concatenating the extracted local and global features.

Similarly, in [16], a model called RAM1 was introduced, extracting local features from a series of local and global regions. This model initially utilizes a CNN-based network to generate a shared feature map. Four deep convolutional models then process this feature map to produce different global and local features. Also, RAM training was complemented by stepwise optimization of the softmax loss in multiple classification tasks that jointly take into account vehicle IDs, types or models, and colors.

Also, J. Zhu et al. in [15] have suggested a quadruple-directed deep learning feature (QD-DLF) to enhance the discrimination ability of global features. Like most previous works, their deep model contains a CNN network with different directional feature pooling layers. A vehicle image is fed into the CNN to obtain its feature map. Then, four moderate-depth (i.e.,16 convolutional layers) QD-DLF networks that sequentially include horizontal, vertical, diagonal, and anti-diagonal average pooling layers have been used to compress the original feature maps into the mentioned directional feature maps. Finally, the final visual appearance feature vector is obtained by spatial normalization and concatenation of these directional feature maps. B. Li et al. in [18] focused on the detrimental effects of interdependence between QD-DLF networks due to the same architecture and adopting the identical training dataset, which reduces the performance of the vehicle ReID model. Their model, called $JQD^3N2$, used an interdependence loss to indirectly decrease the interdependence between QD-DLFs, thereby achieving more distinct visual features for vehicle ReID.

---

[1] Region-Aware deep Model

[2] Joint Quadruple Decorrelation Directional Deep Networks

Similarly, in [17] a Partition and Reunion Network (PRN) has been presented to extract local features with constant partitioning of global feature maps. A ResNet-50 convolutional neural network was utilized as a global feature vector extractor. Then, after the conv4_1 layer in ResNet-50, convolutional layers were duplicated to divide the backbone into height-channel and width-channel branches. Each branch was split into height/ width and channel feature maps. Therefore, these three feature maps, including height, width, and channel feature maps, were generated and connected to obtain the final feature maps. The authors of [52] have developed a two-level attention network consisting of hard part-level and soft pixel-level attention modules to learn more distinctive visual appearance features. The first module reveals parts of the vehicle, such as the windscreen and the head of the car, and the second one concentrates on paying more attention to the distinctive features in each part of the vehicle. Also, they developed a multi-grain ranking loss function that formulates the intra-class compactness and inter-class discrimination objectives to improve the discriminative ability of learned features.

A self-attention model was proposed in the study of C. Liu et al. [20] to extract more subtle characteristics from vehicle images to obtain distinctive features for vehicle ReID. This model consists of a pre-trained ResNet50 up to layer conv4_1 as a backbone network, which continues with four parallel self-attention modules and terminated by a max-pooling layer and a convolutional layer. Four copies of the backbone outputs are fed into four CNN blocks that contain several convolutional layers to extract global features. These global features and some of their modifications are fed as input into ten branches of self-attention to extract multi-level local features. Finally, these features are processed by max-pooling and a convolutional layer to produce the final visual appearance of feature maps. The model is trained by utilizing the cross-entropy loss and triplet loss. Also, some spatiotemporal information of the vehicle movement path is extracted by a Bayesian model to re-rank the model's outcomes.

Similarly, X. Ma et al. [21], inspired by the study of Yu [53], deployed a two-stage attention-based deep model to extract as many discriminative features as possible for vehicle ReID. They adopted an STN3 [54] and a grid generator to automatically isolate vehicles without previous restrictions and split them into three constant parts. These parts were fed as input into the three residual attention models to extract more discriminative appearance visual features.

As mentioned earlier, the part detection approach is another way to learn local features. This approach commonly uses a well-known object detector such as YOLO [55] to find vehicle parts and discover discriminative local features. The main drawback is that the part detector module is normally a deep network and consequently needs a lot of manual annotation, bulk training, and inference calculations. For example, B. He et al. in [56] considered three key vehicle parts, including lights (headlights and taillights), windows (front window and rear window), and vehicle brand. They applied the YOLO to detect these parts in the gallery and probe images. Then, the original image and the three parts were fed into four individual ResNet-50 networks to extract one global feature map and three local feature maps. Finally, the aggregation module fused these feature maps to obtain a distinct feature map for vehicle ReID.

The authors in [31] presented the VAC21 dataset to learn local vehicle attributes and support vehicle ReID models to discover key information from vehicle images. This dataset consists of a gallery with 7129 images of five different types of vehicles, annotated with 21 classes of hierarchical attributes (see Table. I) and their bounding boxes. To the best of our knowledge, this is the only comprehensive dataset in which a wide range of subtle attributes of vehicle images are annotated. Moreover, they trained the single-shot SSD network [57] on this dataset as an attribute detection model for various computer vision tasks such as vehicle ReID. As an illustration, [58] adopted this pre-trained SSD detector to extract vehicle attributes. Only 16 of 21 attributes were selected and fed to a part-guided attention network to identify the areas of the key piece and fuse the extracted local and global feature maps to get more distinct visual features.

TABLE I
NAME AND ABBREVIATION OF 21 CLASSES OF HIERARCHICAL ATTRIBUTES IN THE VAC21 DATASET.

| Name | Abbreviation | Name | Abbreviation |
|---|---|---|---|
| Annual service signs | Anusigns | Logo | Logo |
| Back mirror | backmirror | Newer sign | newersign |
| Bus | bus | Tissue box | paperbox |
| Car | car | Plate | plate |
| Car light | carlight | Safe belt | safebelt |
| Carrier | carrier | Train | train |
| Car top window | cartopwindow | Tricycle | tricar |
| Entry license | entrylicense | Truck | truck |
| Hanging | hungs | Wheel | wheel |
| Lay ornament | layon | Wind-shield glass | windglass |
| Light cover | lightcover | | |

In addition to extracting meaningful parts from vehicle images, some research has focused on automatically localizing several key points from vehicle images and then learning local features from these key points. For example, a two-step approach for key-point approximation is developed in the study of Gu et al. [49]. In the former, the coordination of twenty key points and a $56 \times 56$ heatmap are approximated by adopting a VGG-16 [59] network. In the next step, a double-stack hourglass [60] is used as a refinement network to enhance the heatmaps and decrease the spuriousness due to imperceptible key points. A convolutional network processes these key points and vehicle orientation estimation information to choose the adaptive key points and extract local features. In parallel, a pre-trained ResNet-50 network is used to extract the global features. Finally, the local and global features are concatenated and processed by a fully connected layer to extract the final visual features. Also, in the study of Z. Wang et al. [61], a deep model was introduced to automatically identify vehicle parts and extract orientation-invariant local features in addition to global features for vehicle ReID. More precisely, the positions of twenty vertical key points were estimated using the hourglass network [60] and then clustered by four region proposal masks. These masks were adopted with the original image to discover the global feature vector and four local feature vectors. Finally, these features were combined by a customized convolutional layer to yield the orientation-invariant feature vector. Similarly, in the study of Zheng et al. [62], an image segmentation model based on key points is introduced to partition the raw vehicle image into several foreground parts and detect whether each part is discriminative. A deep network processes a set of discriminative parts plus the original image to extract distinct visual feature maps in the Cartesian domain, where the similarity between any two arbitrary feature maps in this domain can be calculated through the Euclidean distance measure.

---

[3] Spatial Transformer Network

In short, the foremost advantage of constant spatial partitioning is that these methods do not necessitate part labelling and associated time complexity costs but are susceptible to inefficiencies due to mismatching partitions. In contrast, part detection methods can alleviate the mismatch issue but are in contact with the enormous time complexity of manual part labelling and training computations. No matter how to partition the raw image and how to detect parts, both categories learn local features independently on each part region, regardless of the relationship between regions.

- **Considering Relationships Among Part Regions**

These methods are usually developed by combining GCN4s [63] with CNNs to account for the relationships between part regions. CNN typically discovers global features, and GCN is used to learn the relationship among local features calculated from part regions. GCN is a deep neural network that can identify the spatial relations of the graph structure entities. As an illustration, X. Liu et al. [22] introduced PCRNet5 to partition vehicle images into parts, discover part-level distinctive features, and determine the relationships between parts for vehicle ReID. After decomposing the vehicle image into parts through an image segmentation network, the PCRNet adopted two disjoint modules to discover local and global features separately. A CNN-based model was developed to discover global features. A part-neighboring graph was constructed based on the vehicle body structure to consider the associations between parts in local features. Then, a set of GCNs was utilized to propagate local features among the parts and extract the most discriminative local visual features of different viewpoints.

Similarly, HSS-SCN6 inspired by GCN was presented in [23] to understand the hierarchical association between vehicle body parts and to extract more distinctive features for vehicle ReID. Like most previous works, this framework included two modules to provide global and local feature maps. The global features module was implemented through a ResNet-50 network and then fed into the local feature module to form the structural graph network. In the local features module, a constant spatial partitioning method was adopted to divide the global feature map into five local regions including the feature's upper-left, upper-right, middle, down-left, and down-right. These five feature maps and the global feature map form the graph's vertices, and the spatial proximity between all pairs of local or global vertices constitute the edges of the graph. Also, Y. Zhu et al. [24] introduced SGAT7 to consider intrinsic structural associations among markers such as logos, windows, lights, and license plates and extrinsic structural associations among vehicle images. In particular, SGAT contains three elements: appearance, attribute, and extrinsic SGAT (ESGAT) modules. Initially, the appearance module uses a CNN network to extract global features while simultaneously employing intrinsic SGAT (ISGAT) to discover local features. These features are concatenated to form the vehicle's visual appearance features. In parallel, the gallery images are fed into the attribute module to calculate the attribute similarity matrix. Finally, the ESGAT network adopts the similarity matrix to improve the vehicle's visual appearance features. Also, F. Shen et al. [25] proposed HPGN8 by employing pyramidal architecture to combine multiple SGNs to entirely discover the spatial significance of feature maps at different scales. Initially, a ResNet-50 was applied as a backbone network to discover global feature maps of the input vehicle images. Next, multi-scale feature maps were generated by simultaneously resizing the global feature map trough applying five pooling layers. Then, SG9s were constructed at each scale by considering the elements of its corresponding scale feature maps as vertices and the spatial similarity among vertices as edges. At each level of the pyramid architecture, its SGN was created by stacking three SGs to handle the corresponding scale feature maps. The outputs of SGNs were concatenated to produce distinct vehicle visual features. In [64], the authors have adopted a CNN model followed by a transformer [65] to excavate global features and a knowledge graph transfer network, consisting of all vehicle types as nodes to discover inter-class information correlation.

Although methods that consider the relationships among part regions have obtained promising results, they still do not consider the correlation between local features and global features, as well as other descriptive attributes such as color, viewpoint, brand, etc., and as a result, they have not reached enough maturity, and the other efforts have been made by involving transformers to fill these deficiencies.

- **Transformer-based Feature Learning**

The transformer concept was originally introduced in the study of A. Vaswani et al. [65] to handle machine translation issues. Spatial statistics preservation and global processing are two essential opportunities established by transformers. In contrast to CNN models, which lose most of the context-dependent information due to under-sampling operations,

---

transformers preserve spatial statistics and provide faraway information by adopting a multi-head self-attention mechanism.

Afterward, researchers employed transformers to computer vision and concluded remarkable performances in a wide range of vision tasks [27, 66-69] compared to CNNs. For instance, ViT10 has been proposed in the study of A. Dosovitskiy et al. [66] for the image classification problem, and its effectiveness has been confirmed on several well-known benchmarks. Lately, the capabilities of transformers have been adopted to solve vehicle ReID problems. In this section, we have discussed the transformer's supervised and feature learning applications, and its other applications have been reviewed in other sections.

In the study of L. Du et al. [70], ViT schema has been customized for vehicle ReID. The vehicle image has been segmented into patches, linearly projected as local features and merged with viewpoint information to form the inputs to the transformer layers. Similarly, Z. Yu et al. [71] presented VAT11 as a transformer framework to integrate part-level local features and vehicle attributes to achieve more distinct feature maps. The vehicle image has been partitioned into several pieces, linearly projected as visual features, combined with attribute features (color, model, viewpoint, etc.), and fed into transformer layers to produce a vehicle feature map. Also, the multi-sample triblet loss has been adopted to optimize the transformer network.

M. Li et al. [28] focused on the learning challenge arising from salient differences among images of identical vehicles captured from various orientations. They proposed a transformer-based schema to overcome this issue. Their transformer considered the part-level intersections between different orientations by modelling the part-wise correspondence between intra and inter-viewpoints. More precisely, multiple view images were segmented into parts and constricted through the Convnet encoder, and then vehicle representation was extracted by part-level interaction in the transformer.

As mentioned in the previous section, GCN only considers the relationships between part regions and discovers local and global features individually. In the study of F. Shen et al. [26] the GiT12 has been presented, which combines GCN and transformers to extract global and local features and learn the interaction and cooperation between them. In the micro view, the vehicle image is divided into multiple meaning parts called patches, then linearly projected to vectors as vertices to create a local correlation graph (LCG). Finally, the LCG is fed into a transformer layer to build a GiT block. Each GiT block is connected to the next one, models the interaction between local and global features and provides final discriminative feature maps for vehicle ReID.

Some researchers have recently extended the transformer models to use semantic and local visual features to reach more efficient vehicle ReID models. Z. Yu et al. [72] developed the SOFCT13 schema to explore more discriminative global and local features. The global feature extraction pipeline starts by dividing vehicle images into square patches, which are then mapped to the high-dimensional data by a linear projection layer. A token learner is applied to learn this data and is updated during network training to learn the statistical properties of the entire image. These features are combined with other attributes, such as color, position, viewpoint, and model, and fed into the transformer layer to excavate the final global feature vector. Also, the vehicle image is segmented into five classes, including front, rear, top, and side, and accordingly, the image patches are divided into five categories and processed with a transformer layer to extract weighted local features. Similarly, the authors of [73] developed the MART14 framework to efficiently discover foreground-centered global features, extract more discriminative local features, and argue occluded local features. First, to eliminate the background effects on global features, the vehicle image mask is estimated by adopting a U-Net [74] with SEResNeXt50 [75] that predicts the class label (mask value) of each pixel of the vehicle image. Each pixel can receive only one of five class labels, including the vehicle's front, rear, top, and side, digitized by natural numbers from 1 to 4 as its mask value. The vehicle mask is partitioned into overlapping patches and then mapped into the semantic feature domain by a linear projection layer. In parallel, the raw image of the vehicle is directly segmented into several patches and then flatted and transformed into a token feature domain through a linear projection layer. Both the semantic feature map and the token feature map and their position information are concatenated to construct the inputs of the transformer networks to produce foreground-centered global features. In the second step, a directed GCN is constructed for each vehicle image by dividing its corresponding semantic feature map. Then, the adjacency matrix of the GCN is fed into a transformer layer to discover local features. Also, the GCN is utilized to infer local features of a vehicle occluded by other objects. Also, in the study of Z. Li et al. [64], the authors focused on removing background effects in vehicle images without annotations. More precisely, they proposed SMNet, composed of two disjoint modules, NPF and SFE, responsible for background effect reduction and fine-grained feature discovery, respectively. NPF extends ViT as a noise filter that detects the background and removes its effects without annotation. SFE uses a self-attention mechanism to extract the most salient features of the vehicle. Although this model appears to be automated and has very low time complexity, it suffers from insufficient features to achieve high performance compared to similar research.

### 3) Knowledge-based Methods

In the context of vehicle ReID, knowledge refers to any spatio-temporal or textual attributes outside of visual appearance features and vehicle attributes. Spatio-temporal attributes include vehicle trajectories, camera location and neighborhood cameras, weather conditions, daylight status, etc. Knowledge-

---

based methods aim to employ exterior knowledge of visual features and vehicle attributes for vehicle ReID. Generally, no clear boundary can be drawn between these methods and some of the transformer-based feature learning methods mentioned in the previous section that use vehicle attributes and semantic feature domains. However, using spatio-temporal cues is essential for knowledge-based methods to be distinguished from other categories.

Knowledge-based methods have utilized spatio-temporal information to enhance the visual features and refine the retrieved results [5, 12, 32, 48, 76]. For example, Y. Shen et al. [32] defined three parameters, including visual appearance, time stamp, and geographic location of the camera as the visual-spatio-temporal state and introduced a two-stage architecture to consider the visual-spatio-temporal states and efficiently refine the vehicle ReID results. It generates spatiotemporal trajectories for the query and gallery images in the first stage by optimizing a chain MRF[15] model through a deeply learned potential function. An LSTM network then verifies the trajectory, and Siamese-CNN calculates the similarity score to accomplish robust vehicle ReID performance. Similarly, N. Jiang et al. [12] introduced a two-part framework consisting of a CNN-based backbone network to increase the generalization capacity of the vehicle ReID model. The parts aim to extract color, model, and appearance features. Also, a re-ranking technique has been presented to establish spatiotemporal relationships between vehicle images from different cameras and re-rank the analogous appearance retrieved results. As well, J. Peng et al. [77] developed a two-stage schema that employs a multi-task deep network in the first stage to discover distinctive features and adopts a spatiotemporal re-ranking module in the second stage to refine the outcomes of the deep network. In the study of X. Liu et al. [48], the PROVID[16] schema is proposed for vehicle ReID, which considers visual features, license plates, camera locations, contextual information, and the spatio-temporal similarity based on dataset statistics.

Moreover, in [78] spatio-temporal information is utilized to fill the lack of visual features in vehicle ReID tasks. DenseNet121 [79] is a backbone convolutional network to discover visual features and retrieve a set of gallery images for each query. Then, the location and timestamps of the outcome set are used to form a transfer time matrix and filter out outliers and irrelevant gallery images. Similarly, in the study of X. Tan et al. [80], a multi-camera vehicle ReID method has been proposed inside the MCMT task, which takes multi-camera spatio-temporal information to impose some constraints on the outcome images and re-rank them. J. Tu et al. [76] applied a two-branch CNN-based and attention-based module to discover global and local visual features and a spatiotemporal module to set up a distance function to measure location and timestamps similarity between vehicle images. Instead of the transfer time matrix [78] and spatiotemporal constraints [80], the distance function is formulated using random variables distribution, which can be calculated more efficiently and easily extended to large-scale monitoring systems.

However, in some knowledge-based methods, collecting spatio-temporal information needs a large amount of manual annotation work and MCMT tasks, which harms the scalability and generalizability of these methods. Accordingly, in recent works, less attention has been paid to adopting all spatio-temporal information, especially the correlation between the images of neighboring cameras and the movement path of vehicles, and often only a limited number of these attributes have been used. As an illustration, H. Li et al. [5] introduced a transformer-based schema called MsKAT[17], which considers viewpoint and camera position as spatio-temporal information and vehicle color and type as a knowledge vector.

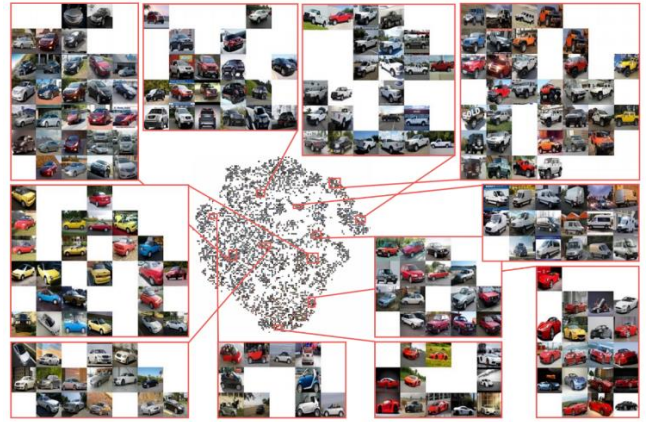

**Fig. 3.** Metric Learning aims to learn a function that maps images into a new space where similarity between objects is preserved (this image was originally prepared by [81]).

### B. Metric Learning

As seen in Figure 3, the key purpose of metric learning is to learn a representation function that maps object images into a novel visual appearance space, where objects with the same class label are as close to each other as possible, and objects with different class labels are further away from each other [82]. Contrastive loss [83] and triplet loss [84] are two fundamental types of loss functions used in metric learning. Furthermore, various loss functions have been developed to customize metric learning for vehicle ReID tasks

Initially, we define some notations. The training set is illustrated as $D = \{I_i \in \mathbb{R}^D | i = 1, ..., N\}$ and $G_w : \mathbb{R}^D \rightarrow \mathbb{R}^d$ denotes a neural network or any parametric function as an embedding function that maps the input sample into an embedding feature space. Given these assumptions, this section overviews loss functions and their applications to the vehicle ReID problem.

- **Cross-entropy loss**

Cross-entropy loss, also called soft-max loss, deals with classification problems and is not included in the metric learning scope. However, due to its application along with metric learning loss functions in vehicle ReID models, it is reviewed in this subsection.

Cross-entropy loss formulates the difference between the

---

target and predicted class labels of each training sample:

$$L(w, I) = -\sum_{i=1}^{M} q_i \log(p_i), \qquad (4)$$

where $p_i$ signifies the prediction possibility that the sample I belongs to class $i$ whose true label is $y$, M denotes the number of classes, and $q_i$ is a binary value calculated as:

$$q_i = \begin{cases} 0, & i \neq y \\ 1, & i = y \end{cases}, \qquad (5)$$

In [85] the LSCE18 loss function has been proposed as follows:

$$L(w, I) = -\sum_{i=1}^{M} \acute{q}_i \log(p_i), \qquad (6)$$

where $\acute{q}_i$ is defined as:

$$\acute{q}_i = (1 - \varepsilon)\delta_{i,y} + \frac{\varepsilon}{K}, \qquad (7)$$

where $\varepsilon$ is a smoothing parameter and $\delta_{k,y}$ is Dirac delta, which equals 1 for $i = y$ and 0 otherwise.

- **Contrastive loss**

Contrastive loss takes a pair of training points marked similar or dissimilar and maps them into an embedding space, making the similar points closer and the dissimilar points far apart. Formally, for each arbitrary pair of training points $I_i, I_j \in \mathbb{R}^D$, a binary label is assigned as follows [83]:

$$y_{i,j} = \begin{cases} 0, & if\ I_i, I_j\ marked\ as\ similar \\ 1, & otherwise \end{cases}, \qquad (8)$$

then the contrastive loss is defined as follows:

$$\mathcal{L}(w) = \sum_{i=1}^{N} \sum_{j=1}^{N} L(w; I_i, I_j, y_{i,j}), \qquad (9)$$

where:

$$L(w; I_i, I_j, y_{i,j}) = \frac{1}{2}(1 - y_{i,j})\|G_w(I_i) - G_w(I_j)\|^2 + \frac{1}{2}y_{i,j}\max\{0, m - \|G_w(I_i) - G_w(I_j)\|^2\}, \qquad (10)$$

and $m > 0$ is a hyperparameter clarifying the minimum distance between dissimilar points.

- **Triplet loss**

Google initially presented triplet loss in FaceNet [84] for the face recognition task. Theoretically, each vehicle image in the training set is a positive sample. $I_i^p$, an image with the same identity is randomly selected as an anchor sample $I_i^a$, and other images with different identities are considered as negative samples $I_i^n$, and Triplet loss attempts to guarantee that the anchor sample is closer to a positive sample than any other negative sample. More formally, the Triplet loss is defined as follows:

$$L(w) = \sum_{i=1}^{N} \left( \|G_w(I_i^a) - G_w(I_i^p)\|^2 - \|G_w(I_i^a) - G_w(I_i^n)\|^2 + \alpha \right), \qquad (11)$$

where $\alpha$ is a hyperparameter representing the minimum distance between dissimilar samples. Due to the simultaneous use of positive and negative samples while training the network using Triplet loss, fewer training samples are required for convergence compared to Contrastive loss.

Recently, several researchers have been reported that have utilized Triplet loss and its variations for vehicle ReID tasks [86]. In the study of Y. Tang et al. [87], a metric learning schema was introduced to combine deep features and vehicle attributes, where a triplet loss function is applied to optimize the model. In [88], triplet loss is adopted to develop a viewpoint-aware metric learning methodology. More precisely, the sum of three independent triplet loss functions for similar, dissimilar and cross viewpoints is considered a total loss and jointly optimized to handle matching vehicle images in a camera with different viewpoints. Also, to achieve a viewpoint-invariant embedding function with low time complexity, the authors of [89], [90] and [91] have proposed to employ BH19 as hardest, BA20 as uniformly, BS21 as multinomial and BW22 as adaptive weights sampling methods in the triplet loss function. To be more precise, let $P(I^a)$ and $N(I^a)$ be respectively a subset of positive and negative instances analogous to an anchor instance $I^a$. Then, the modified triplet loss can be defined as follows:

$$L(w, I^a) = \sum_{\forall I^p \in P(I^a)} w_p \|G_w(I^a) - G_w(I^p)\|^2 - \sum_{\forall I^n \in N(I^a)} w_n \|G_w(I^a) - G_w(I^n)\|^2 + \alpha, \qquad (12)$$

where $w_p$ and $w_n$ denotes the weights of positive and negative instances, respectively. The total loss is also defined as:

$$L(w) = \sum_{\forall I^a \in D} L(I^a) \qquad (13)$$

Also, an evaluation of these different sampling methods on the vehicle ReID problem is reported in the study of R. Kumar et al. [92], confirming the modified triplet loss's effectiveness compared to the traditional triplet loss. Similarly, to differentiate between large and small absolute distances between positive and negative samples, an improvement on the hard sampling triplet loss has been presented in [93] by adding a balancing term to its function.

J. Yu et al. [94] proposed the DTL23 function as a modification of Triplet loss for self-supervised metric learning. Specifically, they proposed an unsupervised vehicle ReID model that builds a feature dictionary from vehicle images and employs the DTL to process the dictionary, train the model with unlabeled data, and enhance the distinctiveness of the learned features. Y. Bai et al. in [95] brought up low convergence speed and insufficient feature discrimination power as two disadvantages of triplet loss and introduced ICV24 triplet loss to overcome these issues. They adopted a multi-task learning strategy and jointly optimized the ICV triplet and cross-entropy loss to generate more discriminative features for vehicle ReID. Also, VAL [96] and VARID [97] have been proposed as two viewpoint-aware triplet loss functions to solve the harmful effects of intra-class dissimilarity and inter-class similarity problems in vehicle ReID. In particular, the intra-view triplet loss function has been defined to take into account the discrimination of different vehicles from the same viewpoint, and the inter-view triplet loss function has been deployed to impose the same vehicle samples in different viewpoints to be close to each other [97]. The viewpoint-aware loss function is calculated as the weighted average of inter-view and intra-view loss functions.

---

[18] Label-Smoothing Cross Entropy
[19] Batch hard
[20] Batch all
[21] Batch sample

[22] Batch weighted
[23] Dictionary-based Triplet Loss
[24] Intra-Class Variance

Several recent works have focused on combining classification and metric learning to overcome the frailties of triplet loss [33, 98]. In [98], [86] and [99], various combinations of triplet loss and cross-entropy loss are adopted as the optimization function during the training of a deep network. In [33], the logistic loss function is presented as an extension of the triplet loss and applied with the LSCE loss function [85] to train the vehicle ReID model. The purpose of logistic triplet loss is to compensate the information ignoring issue due to the margin hyperparameter $\alpha$ and is defined as follows:

$$L(w) = \sum_{i=1}^{N}(\log(1 + e^{z_i})), \tag{14}$$

where $z_i = \left\| G_w(I_i^a) - G_w(I_i^p) \right\|^2 - \left\| G_w(I_i^a) - G_w(I_i^n) \right\|^2$. Also, the authors of [100] claimed that the diversity of traffic scenes in real-world environments causes the class imbalance problem between positive and negative samples. To address this issue, they adopted the combination of Triplet loss with Focal loss [101] and Range loss [102].

- **Large-margin loss**

LML25 [103] assumes that a CNN-based network has already been trained by an arbitrary classification loss function (i.e., entropy loss) and an embedding feature space is obtained, and attempts to fine-tune the network. Inspired by SVM26, the LML looks for decision boundaries in the embedding space to separate instances with different identities. It specifically places hard instances close to the center of their own class region in the embedding space via network propagation to solve the intra-class dissimilarity issue.

More precisely, for training instances with the same class label as positive samples, and other training samples as negative samples, a large margin decision boundary $w^T . x + b = 0$ is obtained. Then, the distanced between positive instances and their corresponding decision boundary is calculated as follows:

$$d = \sum_{\forall x_i \in positive\ batch} \frac{y_i(w.x_i+b)}{\|w\|} \tag{15}$$

and the LML term is defined as:

$$L_{LML} = e^{-d/\sigma}, \tag{16}$$

where $\sigma$ is a parameter to control the distance distribution, finally, LML function is defined by the combination of cross-entropy and LML term as follows:

$$L = L_{cross-entropy} + \lambda L_{LML}, \tag{17}$$

where $\lambda$ is a regularization parameter. Also, a back propagative LML-based metric learning algorithm is presented in [103] to solve the vehicle ReID problem.

- **Group-Group Learning Loss**

GGL27 [8] has been proposed to refine the difficulties of instance selection sensitivity and low convergence speed challenges in triplet loss [84, 104]. Initially, the training data set is partitioned into groups, each containing all the images of only one vehicle identity. GGL then updates the model weights to compact the instances with the same identity close to each other in the embedding feature space and to place different groups as far apart as possible.

Let $M$ denotes the number of identities, $n^i$ represents the number of instances of $i^{th}$ identity, and the feature vector of $j^{th}$ instance of $i^{th}$ identity is denoted by $f_j^i$ for $i = 1, 2, \ldots, M$ and $j = 1, 2, \ldots, n^i$. In the embedding feature space, the mean and variance of samples of $i^{th}$ identity are defined as follows:

$$m^i = \frac{1}{n^i}\sum_{j=1}^{n^i} f_j^i \; ; \qquad var^i = \frac{1}{n^i}\sum_{j=1}^{n^i}\left\| f_j^i - m^i \right\|^2, \tag{18}$$

for $i = 1, 2, \ldots, M$. Following these assumptions, the GGL loss function is defined as follows:

$$\mathcal{L}_{GGL} = \mathcal{L}_{intra} + \lambda\mathcal{L}_{inter}, \tag{19}$$

where $\lambda$ is a weighting control parameter, $\mathcal{L}_{intra}$ and $\mathcal{L}_{inter}$ are intra-group and inter-group loss functions, respectively, and are defined as follows:

$$\mathcal{L}_{intra} = \frac{1}{M}\sum_{i=1}^{M} var^i, \tag{20}$$

$$\mathcal{L}_{inter} = \frac{1}{M(M-1)}\sum_{i,j=1;\ i\neq j}^{M} \max\{0, \alpha - \frac{1}{2}\left\| m^i - m^j \right\|^2\}, \tag{21}$$

where $0 \leq \alpha \leq 1$ is margin parameter to enhance the effectiveness of network learning.

- **Other Loss Functions**

In addition to the loss functions studied in the previous sections, various loss functions have been adopted in various computer vision tasks but, to the best of our knowledge, have not been applied to the vehicle ReID problem.

T. Lin et al. [101] developed focal loss to treat the deficiency of cross-entropy loss in the face of the class imbalance problem that often arises in training dense object detectors. Also, range loss function [102] and circle loss [105] have been introduced to decline intra-class dissimilarity while increasing inter-class similarity. Similarly, the center loss has been deployed to improve the discriminative power of features in face recognition tasks [34]. Multi-Grain Ranking loss has been introduced to achieve the most discriminative deep features in vehicle ReID [52]. Multi-class N-pair loss has been introduced as a generalization of traditional triplets to overcome the slow

---

convergence issue [106]. Also, in the study of X. Wang et al. [104], the ranked-list loss has been proposed to achieve fast convergence and high performance in metric learning, and in the study of E. Kamenou et al. [34], it has been utilized for vehicle ReID task.

| Transfer Learning | Target Domain | |
|---|---|---|
| | Labeled | Unlabeled |
| Source Domain · Labeled | Multi-task Learning | Unsupervised Domain Adaptation |
| Source Domain · Unlabeled | Self-taught Learning | Unsupervised Learning |

**Fig. 4.** Types of unsupervised learning based on transfer learning categories.

## IV. UNSUPERVISED VEHICLE RE-IDENTIFICATION

The rapid growth of CNN-based deep learning technologies has led to various supervised vehicle ReID methods in recent years. The performance of the supervised method mostly depends on the availability of a large-scale domain-specific annotated training set, which requires a costly and time-consuming task to prepare. In particular, the insufficient training set causes the efficiency to fall exponentially whenever a supervised model is transferred to real-world large-scale surveillance systems. From the academicians' point of view, unsupervised learning is an effective approach to overcome these restrictions, capturing the most meaningful patterns from a dataset without annotating any training set.

Deep-learning-based unsupervised methods have usually been developed based on transfer learning, where a model trained on one data set as a source domain is refined or adapted to work on a different but related data set as the target domain. Depending on whether the source and target domains are already labeled or not, transfer learning can be divided into four categories, as shown in Fig. 4. As a result of this classification, unsupervised vehicle ReID techniques can be categorized into two main groups: unsupervised domain adaptation and fully unsupervised methods. This section offers a comprehensive overview of these approaches.

### A. Unsupervised domain adaption methods

Unsupervised domain adaptation involves training a model on data from one domain (source domain) and adapting it to perform on another domain (target domain) without having any labeled target domain data during training. The goal is to reduce the disparity between the source and target domains, enabling the model to generalize better when faced with new and unseen data. Several approaches have been proposed for unsupervised domain adaptation in the machine learning scope, including domain adversarial training, instance-based methods, self-ensembling techniques, feature alignment, and GAN-based methods. These methods employ various strategies to address domain discrepancy during model training. For instance, domain adversarial training, like DANN [107], utilizes a domain discriminator to align representations from different domains while training a primary task. Another approach, instance-based methods, aims to match domain distributions by aligning instances or features using techniques like MMD or CORAL [121]. Self-ensembling techniques, such as Mean Teacher [108], VAT [109], PSUReID [110] and HyPASS [111], employ consistency regularization to stabilize predictions across diverse domain shifts through pseudo-labeling. Feature Alignment techniques, such as Deep CORAL [112] and VAE [113], modify model architectures or learning objectives to explicitly align features across domains, often incorporating domain-specific normalization or adaptation layers. GAN-based methods, exemplified by CycleGAN [37] and DiscoGAN [114], extend the GAN framework to generate synthetic target domain data resembling the source domain, effectively reducing domain differences. To the best of our knowledge, just a few studies about unsupervised vehicle ReID especially unsupervised domain adaptation, have been done, and most of the research has focused on person ReID task.

The authors in [115] and [116] addressed the challenge of significant performance degradation caused by distribution disparities between training (source) and test (target) datasets. The challenge arises from the heterogeneous nature of distinct domains, manifesting in diverse image characteristics, including varying backgrounds, illuminations, resolutions, and discrepancies in camera perspectives. To overcome this challenge, J. Peng et al. [115] introduced VTGAN , an image-to-image translation framework aimed at transferring the style of the source domain onto the target domain while preserving their identity information. Also, they introduced ATTNet , leveraging an attention-based structure to train generated images, thereby enabling the discovery of more distinctive features while suppressing background in the context of vehicle ReID.

In the study of C.-S. Hu et al. [117], the transformation of vehicle pose has been articulated as a domain adaptation task. PTGAN has been designed to receive key points representing the viewpoints of vehicles and subsequently generate a fake image corresponding to a novel viewpoint to tackle the pose variation problem. Also, the task of identifying identical vehicles across diverse domains, i.e., encompassing both day-time and night-time domains, has been explored in [118] as a domain adaptation issue. A GAN-based framework has been introduced to transform domains for two input images into an image belonging to an alternative domain. Subsequently, a four-branch Siamese network was utilized to learn the distance metric between images belonging to two distinct domains.

Representations learned through UDA typically lack task-specific orientation, meaning they often do not simultaneously possess characteristics of being class-discriminative and domain-transferable. Efforts have been dedicated to addressing this concern within UDA for vehicle ReID. Notably, authors in [119] introduced DTDN. This framework disentangles data representations into two distinct components: one comprising task-relevant elements, encapsulating vital cross-domain task-related information, and the other encompassing task-irrelevant facets, containing non-transferable or disruptive data. Task-specific objective functions have been employed across domains to regulate these components. This regularization explicitly promotes disentanglement without necessitating the utilization of generative models or decoders. An initial exploration of employing a Transformer in UDA vehicle Re-ID was introduced in the study of R. Wei et al. [116] to overcome the mentioned issue. This Transformer-based network aimed to enhance the integration of contextual information within images. Specifically, the network adaptively directed attention to discriminative vehicle components across the source and target domains. It incorporated a domain encoder module to discern domain-invariant characteristics and alleviate the impact of domain-related factors. Moreover, a contrastive clustering loss was applied to perform clustering on the feature representations of target samples before each training epoch. These clusters were subsequently assigned labels, functioning as pseudo-identities to supervise the subsequent training procedure.

Pseudo-labeling techniques remain the dominant choice for addressing UDA vehicle ReID tasks due to their superior performance [29, 110]. Nevertheless, the efficacy of pseudo-labeling hinges significantly upon selecting specific hyperparameters that directly influence the generation of pseudo-labels through clustering methods. Addressing this challenge, the approach outlined in [111], known as HyPASS, is introduced as a technique tailored for the automated and cyclic adjustment of hyperparameters in pseudo-labeling for UDA clustering. HyPASS comprises two fundamental components within the framework of pseudo-labeling methods: initially, the selection of hyperparameters is contingent upon a validated set derived from labeled source data; subsequently, the refinement of hyperparameter selection is facilitated through conditional alignment of feature discriminativeness, a process honed through the analysis of source samples. Likewise, the study of Z. Lu et al. [29] introduces MAPLD, a method that enhances the precision of pseudo-labeling techniques while effectively mitigating pseudo-label noise within the context of UDA for vehicle ReID.

Despite the accomplishments of UDA-based approaches in vehicle ReID, they often require additional information sourced from diverse domains, potentially limiting their applicability in real-world scenarios. As a result, sometimes fully unsupervised methods are favored due to their compatibility with practical applications, avoiding the need for supplementary data from different domains.

### B. Fully unsupervised methods

Fully unsupervised methods can derive meaningful information directly from unlabeled data, eliminating the need for annotations or labeled data. This characteristic renders these methods more suitable and adaptable for real-world applications and scenarios. These methods predominantly concentrate on developing diverse clustering techniques and progressive training strategies as their key focus areas.

Progressive learning follows a step-by-step approach to learning information, starting from simpler concepts and advancing towards more complex ones. This approach has found extensive application across different computer vision tasks like face recognition [120], classifying images [121], person ReID [122], etc. DUPL-VR [38] and VR-PROUD [39] were specifically crafted to address the challenges associated with vehicle ReID by implementing a progressive strategy in an unsupervised manner. Initially, unlabeled images are inputted into a foundational CNN network, utilizing pre-established weights to extract features. These features then undergo a clustering process, receiving cluster IDs that function as "pseudo" labels for subsequent iterations. Specific heuristic constraints are applied to refine the clustering outcomes to bolster the accuracy and stability of the clusters. The resultant clusters, representing vehicles, are then used to fine-tune an additional CNN network, mirroring the architecture of the foundational CNN. This iterative procedure continually enlarges the training dataset by incorporating increasingly robust clusters, enabling unsupervised self-progressive learning until convergence. A viewpoint-aware clustering approach for unsupervised vehicle Re-ID centered on progressive learning was introduced by A. Zheng et al [123]. Initially, the process involved extracting viewpoint details using a viewpoint prediction network, while simultaneously learning distinctive features for each sample through the utilization of the repelled loss function. Subsequently, the feature space was partitioned into distinct subspaces based on the anticipated viewpoints. This was followed by applying a progressive clustering algorithm to discover precise relationships among the samples and enhance the discriminative ability of the network.

Similarly, the researchers referenced in [3] applied progressive learning to address unsupervised vehicle ReID challenges. Inspired by [124], their primary focus lies in discerning reliable samples and implementing a progressive algorithm for training the network. In particular, their approach had two essential differences: initially, they engineered a multi-branch backbone to capture global and local features, leveraging this dual information to create reliable clusters and thereby mitigating the influence of hard samples. Additionally, their methodology commenced with separate utilization of global and local features for training in the initial stage, gradually transitioning to the fusion of these features as the network's capabilities evolved in subsequent stages.

## V. DATA SETS

Different research groups have prepared many standard datasets and benchmarks to validate the superiority of vehicle ReID models. This section will examine these datasets in detail, particularly their advantages and limitations.

### A. VehicleID [10]

Numerous nonoverlapping surveillance cameras collected the "VehicleID" dataset in a small-scale city in China during the daytime, and there are an average of 8.44 images per vehicle (altogether 221763 images from 26267 vehicles).

In this data set, among 90196 images, 10319 vehicles have been annotated with their model information (only 250 most popular models). Additionally, two orientations of the vehicle, including front or rear, are considered, and view information is not annotated. Every vehicle contains more than one image; subsequently, the dataset is appropriate for vehicle retrieval tasks.

This data set includes a training set and a test set. The training set involves 110178 images of 13134 vehicles, of which 47,558 are tagged with vehicle model information, and the test set includes 111585 images of 13133 vehicles, of which 42638 images are tagged with vehicle model information.

This data set has been gathered in relatively constrained situations with about 20 cameras during daylight, including two orientation views, a few illumination changes and a simple background. Therefore, it does not contain the required benchmarks to assess all the scenarios and challenges of vehicle ReID.

This data set was collected in relatively limited situations involving about 20 cameras in daylight, including two orientations, with low illumination variations, and with a simple background. Therefore, it does not include the required benchmarks and data to evaluate all vehicle ReID challenges.

### TABLE II
NAME AND ABBREVIATION OF 21 CLASSES OF HIERARCHICAL ATTRIBUTES IN THE VAC21 DATASET.

| Data Set | | no. of vehicles | no. of models | no. of colors | no. of images |
|---|---|---|---|---|---|
| VD1 | Train | 70591 | 1232 | 11 | 422326 |
| | Test | 71165 | | | 424032 |
| VD2 | Train | 39619 | 1112 | 11 | 342608 |
| | Test | 40144 | | | 347910 |

### B. VeRI-776 [125]

The VeRi-776 dataset was constructed from the VeRi dataset in [126]. About 20 traffic surveillance cameras collected the VeRi dataset under diverse conditions, such as orientations, illuminations, and occlusions. It contains 40,000 images of 619 vehicles, annotated with various attributes, including vehicle bounding boxes, brands, types, and colors.

The VeRi-776 dataset was produced by expanding VeRi in three aspects: increasing the data size, considering the license plate, and considering the vehicle trajectory as spatio-temporal information. It contains over 50000 vehicle images, 776 vehicle identities, and approximately 9000 trajectories. The VeRi-776 dataset involves a training set of 576 vehicles and 37781 images and a testing set of 200 vehicles and 11579 images.

This dataset was recorded from 4:00 PM to 5:00 PM on a single day and in a round path over a small area of $1\ km^2$ and hence suffers from the lack of sufficient test bed to evaluate all challenges of vehicle ReID.

### C. VD1 and VD2 [45]

The VD1 and VD2 datasets are collected from the front view of vehicles through traffic cameras and surveillance videos, respectively. The vehicle color, model, and identification number are provided as attribute vectors for each vehicle image in both datasets.

The VD1 comprises 846358 images of 141756 vehicles with

11 colors and 1232 models, divided into training and testing sets. VD2 has 807260 images of 79763 vehicles with 11 colors and 1112 models. The characteristics of the training set and the testing set of both datasets are shown in Table. II.

These datasets are greatly simplified in vehicle ReID challenges, as most images are captured from a single view. As a result, the performance on these datasets is already saturated, and a recent method achieved 97.8 95.5% accuracy on VD1 and VD2, respectively.

### TABLE III
THE 5 SCENARIOS OF CITYFLOW DATASET FOR MTMC VEHICLE TRACKING [1].

| # | Time (min.) | no. of cameras | no. of BB | no. of ID | Scene type |
|---|---|---|---|---|---|
| 1 | 17.13 | 5 | 20772 | 95 | Highway |
| 2 | 13.52 | 4 | 20956 | 145 | Highway |
| 3 | 23.33 | 6 | 6174 | 18 | Residential |
| 4 | 17.97 | 25 | 17302 | 71 | Residential |
| 5 | 123.08 | 19 | 164476 | 337 | Residential |

### D. VRIC [14]

This dataset contains 60,430 images of 5,622 vehicle identities collected by a surveillance system with 60 cameras of complex road traffic during the day and night. The VRIC consists of 24 monitoring locations and covers an almost unlimited appearance of vehicles collected by differences in resolution, motion blur, weather conditions, and occlusion.

This data set is divided into training and test sets. The training set includes 54,808 images of 2,811 identities, and the test set includes 5,622 images of 2,811 identities. All images were annotated with car model, color and type.

Although most real-world challenges of vehicle ReID are considered in VRIC dataset capture, it suffers from limited vehicle types and models, lack of detailed vehicle attributes, small-scale imaging area, and overlapping of training set and test set cameras.

### E. CityFlow [1]

Forty non-overlapping surveillance cameras collected the "CityFlow" dataset with a maximum distance of 2.5 km between any two cameras spanning ten intersections in a medium-sized city in the USA. It contains 229680 vehicle images of 666 different vehicle identities, each moving through at least two cameras.

The dataset comprises 3.25 hours of video and contains locations such as highways, roads and intersections. The offset of starting time for each video is annotated, and most videos were captured at ten frames per second with a resolution of at least 960p.

The CityFlow is the first public benchmark that supports multi-target multi-camera (MTMC) vehicle tracking. For this purpose, five scenarios are shown in Table. III.

Also, a subset of CityFlow, called CityFlowReID, has been developed for image-based vehicle ReID. The CityFlow-ReID consists of a total of 56277 vehicle images and 666 different vehicle identities from 4.55 camera views on average. It involves a training set of 333 vehicle identities and 36935 images and a testing set of 333 vehicles and 18290 images.

TABLE IV
A COMPARISON OF LARGE-SCALE DATASETS PRESENTED FOR VEHICLE REID MODEL EVALUATION.

| Measures | Data Sets | | | | | |
| --- | --- | --- | --- | --- | --- | --- |
| | VehicleID | VD1 | VD2 | VeRI-776 | CityFlow | VERI-Wild 2.0 |
| Images | 221763 | 846358 | 807260 | 49360 | 229680 | 825042 |
| Identities | 26267 | 71165 | 40144 | 776 | 666 | 42790 |
| Training samples | 110178 | 422326 | 342608 | 37781 | 36935 | 277797 |
| Test samples | 111585 | 424032 | 347910 | 11579 | 18290 | 398728 |
| Cameras | 2 | N/A | N/A | 20 | 40 | 274 |
| Images/identities | 8.4 | 6 | 10.1 | 63.6 | 84.5 | 59 |
| Capturing area | N/A | N/A | N/A | $1\ km^2$ | Less than $6.25\ km^2$ | $200\ km^2$ |
| Time stamp | N/A | N/A | N/A | Yes | Yes | Yes |
| Vehicle colors | N/A | 11 | 11 | 10 | 10 | 12 |
| Vehicle types | N/A | 1232 | 1112 | 9 | 9 | 13 |
| Vehicle brands | N/A | | | 30 | N/A | 153 |
| Occlusion | No | No | No | Yes | Yes | Yes |
| Complex background | No | No | No | Yes | Yes | Yes |
| Morning | Yes | Yes | Yes | No | Yes | Yes |
| Afternoon | Yes | Yes | Yes | Yes | Yes | Yes |
| Night | No | No | No | No | No | Yes |
| Rainy | No | No | No | No | No | Yes |
| Foggy | No | No | No | No | No | Yes |
| MTMC | No | No | No | No | Yes | No |
| Multi-view? | No | No | No | No | Yes | Yes |
| Training and test camera overlap? | Yes | Yes | Yes | Yes | Yes | No |
| Camera Geometry available? | N/A | N/A | N/A | Yes | Yes | Yes |

*F. VERI-Wild 2.0 [127]*

The VERI-Wild 2.0 dataset was developed by extending VERI-Wild [128] to comprehensively evaluate the discrimination and generalization capabilities of vehicle ReID models. This dataset was collected around the clock for one month by 274 cameras from a huge CCTV surveillance system in an urban area of over $200\ km^2$.

This dataset contains many vehicles of the same model, whose samples include very complex backgrounds, various orientations, severe occlusion, and different weather conditions. It contains 825,042 images of 42,790 vehicle identities, covering various scenarios, including urban roads, street traffic light areas, intersections, highway tolls and ramp entrances. More than 30 per cent of identities were recorded during daytime and nighttime. Also, auxiliary attributes, vehicle color, vehicle types, and vehicle brands were annotated to improve the vehicle's visual feature map.

This dataset has many instances with the same vehicle brand or model. On average, about 59 samples of each vehicle identity were taken from different viewpoints in foggy, rainy and sunny weather conditions. More precisely, 7.1 and 3.48 percent of the data were captured on foggy and rainy days, respectively.

This dataset is divided into training and test sets containing 277797 and 398728 images, respectively. Moreover, 174 cameras capture the training set, and 100 cameras are utilized for the test set. The nonoverlapping of the training and test cameras has made it possible to evaluate the discrimination and generalization capabilities of vehicle ReID models in different light conditions, viewpoints and complex backgrounds. The test set is divided into three subsets of images to fully evaluate the vehicle ReID models and investigate the effect of orientation, lighting and weather changes.

A comprehensive comparison between the investigated datasets for vehicle ReID based on different criteria is shown in Table. IV.

## VI. EVALUATION STRATEGIES

Numerous assessments outlined in the existing literature highlight that dataset selection and performance metrics are key aspects of vehicle ReID's evaluation strategies. By employing these evaluation strategies, researchers and practitioners can comprehensively assess and compare the effectiveness of different vehicle ReID approaches, paving the way for advancements in this domain.

### A. Dataset Selection

Employing suitable standard datasets comprising various vehicle images obtained across varying scenarios, temporal settings, and environmental conditions is crucial for evaluating research challenges. Implementing stringent training and testing divisions within datasets is vital to ensure impartial assessment and to avoid overfitting issues. Also, the generalizability and robustness of the models can be measured by evaluating them in different data sets. This validates the model's performance in various settings beyond the training data. Additionally, it is essential to consider various evaluation scenarios such as single-shot or multi-shot re-identification, intra-camera or inter-camera matching, and conducting evaluations under different conditions such as occlusions or variations in lighting.

The preceding section provided a comprehensive review of established datasets curated for assessing vehicle ReID models.

### B. Performance Metrics

Metrics such as Rank, mAP[1], precision-recall curves and CMC[2] curves are commonly used to measure the effectiveness of re-identification algorithms.

- **Rank Ratio**

In vehicle ReID, the process of ranking vehicles entails thoroughly comparing the features extracted from a query vehicle image with those of all images in the gallery. This comparison results in the images being arranged in descending order of similarity, positioning the most similar images at the higher ranks of the list. Regarding the ground truth of the selected dataset, the first position within the ordered list, where its image corresponds to the same vehicle as the query image, illustrates its rank and holds paramount importance in the evaluating models.

The Rank Ratio criterion is derived by dividing the count of accurately re-identified query images within a specified rank by the total number of query images in the test set. This metric directly indicates the system's efficacy in precisely re-identifying vehicles within a predefined rank. For example, the Rank Ratio at Rank-k shows the algorithm's accuracy in correctly identifying vehicle images up to the k-th rank among the retrieved results for each query.

- **mAP**

Precision and recall, while valuable as single-value criteria, are based on the entire set of images retrieved by the ReID scheme. In the context of vehicle ReID, where models provide a ranked list of images, it is beneficial to consider each image's position in the ranked list. The precision-recall curve can be achieved by calculating precision and recall at each position within this list. The Average Precision (AP) is then defined as follows:

$$AP = \frac{\sum_{k=1}^{M} p(k) \times gt(k)}{N_{gt}}, \qquad (22)$$

where $M$ represents the length of the ordered list, $N_{gt}$ stands for the number of images sharing the same identity as the query within this ordered list, $p(k)$ denotes the precision at the specified cut-off rank $k$, and:

$$gt(k) = \begin{cases} 1 & \text{if label of } k^{th} \text{ img is same as query} \\ 0 & \text{otherwise} \end{cases} \qquad (23)$$

Finally, for a set of queries in a test set, the mAP metric is the mean of the AP scores for each query and can be calculated as follows:

$$mAP = \frac{\sum_{q=1}^{Q} AP(q)}{Q}, \qquad (24)$$

where $Q$ is the number of queries.

- **CMC**

The CMC is a criterion to evaluate how well a vehicle ReID model performs in identifying a specific vehicle among a ranked results list. More precisely, in the context of vehicle re-identification, CMC@k stands for CMC at rank k. It is a metric used to evaluate the performance of a re-identification system by measuring how accurately it retrieves the correct vehicle match within the top k ranks.

By plotting the CMC curve, we can understand how likely it is to find the correct match as we move through different top positions in the ordered list of retrieved results. For instance, if the CMC curve shows 95% accuracy at position 3, there is a 95% chance of finding the correct match among the top 3 positions in the list.

---

[1] mean Average Precision

[2] Cumulative Matching Characteristic

TABLE V

PERFORMANCE OF STATE-OF-THE-ART APPROACHES IN VEHICLE REID ON VERI-776 AND VEHICLEID DATASETS.

| Model | VeRI-776 | | | VehicleID | | | | | | Category |
|---|---|---|---|---|---|---|---|---|---|---|
| | | | | Small | | Medium | | Large | | |
| | mAP | R1 | R5 | R1 | R5 | R1 | R5 | R1 | R5 | |
| OIFE [61] | 48.0 | 65.9 | - | - | - | - | - | 67.0 | 82.9 | Learning of global features |
| MAD [12] | 61.1 | 89.3 | 94.8 | - | - | - | - | - | - | |
| AAVER [49] | 61.2 | 89.0 | 94.7 | 74.7 | 93.8 | 68.6 | 90.0 | 63.5 | 85.6 | |
| FACT [129] | 18.8 | 52.2 | 72.9 | 49.5 | 68.0 | 44.6 | 64.2 | 39.9 | 60.2 | |
| VGG+C+T+S [44] | 57.4 | 86.6 | 92.9 | 69.9 | 66.2 | 63.2 | 87.3 | 82.3 | 79.4 | |
| PROVID [48] | 48.5 | 76.9 | 91.4 | 48.9 | 69.5 | 43.6 | 65.3 | 38.6 | 60.7 | Combining Global Features and Local Features |
| QD-DLF [15] | 61.8 | 88.5 | 94.5 | 72.3 | 92.5 | 70.7 | 88.9 | 64.1 | 83.4 | |
| UMTS [130] | 75.9 | 95.8 | - | 80.9 | - | 78.8 | - | 76.1 | - | |
| RAM [16] | 61.5 | 88.6 | 94.0 | 75.2 | 91.5 | 72.3 | 87.0 | 67.7 | 84.5 | |
| TBENet [131] | 79.5 | 96.0 | 98.5 | - | - | - | - | - | - | |
| PRN [56] | 74.3 | 94.3 | 98.9 | 78.4 | 92.3 | 75.0 | 88.3 | 74.2 | 86.4 | |
| HPGN [25] | 80.2 | 96.7 | - | 83.9 | - | 80.0 | - | 77.3 | - | |
| SAVER [132] | 79.6 | 96.4 | 98.6 | 79.9 | 95.2 | 77.6 | 91.1 | 75.3 | 88.3 | |
| PVEN [133] | 79.5 | 95.6 | 98.4 | 84.7 | 97.0 | 80.6 | 94.5 | 77.8 | 92.0 | |
| PPT [134] | 80.6 | 96.5 | 98.3 | 79.6 | 92.3 | 76.0 | 89.4 | 74.8 | 87.0 | |
| ViT | 78.9 | 95.8 | - | 82.4 | 87.1 | 76.4 | 83.2 | 73.3 | 80.1 | Transformer-based Feature Learning |
| GiT | 80.3 | 96.9 | - | 84.7 | - | 50.5 | - | 77.9 | - | |
| Vit-ReID | 82.1 | 97.4 | - | 85.2 | 97.5 | - | - | - | - | |
| MART | 82.7 | 86.6 | 98.7 | 94.2 | 98.6 | 93.9 | 97.5 | 90.2 | 96.2 | |
| SOFCT | 80.7 | 96.6 | 98.8 | 84.5 | - | 80.9 | - | 78.7 | - | |
| MVPWC [28] | 86.0 | 80.9 | 96.2 | 86.8 | 97.3 | 83.4 | 95.8 | 80.6 | 93.1 | |
| SFMNet | 80.0 | 97.0 | - | 85.1 | 97.1 | 80.5 | 94.7 | 77.6 | 92.5 | |
| MsKAT [5] | 82.0 | 97.1 | 99.0 | 86.3 | 97.4 | 81.8 | 95.5 | 79.4 | 93.9 | Knowledge-based Features |
| DRF-ST [76] | 84.5 | 93.0 | 97.1 | 82.7 | 95.0 | 77.7 | 91.3 | 77.0 | 88.5 | |
| VARID [97] | 79.3 | 99.2 | 96.0 | 85.8 | 96.9 | 81.2 | 94.1 | 79.5 | 92.2 | Metric Learning |
| VAL [96] | 81.4 | 96.7 | - | 85.7 | 97.0 | 81.3 | 95.0 | 78.2 | 93.0 | |
| GRF+GGL [8] | 61.7 | 89.4 | 95.0 | 77.1 | 92.8 | 72.7 | 89.2 | 70.0 | 97.1 | |
| TAMR [52] | - | - | - | 66.0 | 79.7 | 62.9 | 76.8 | 59.7 | 73.9 | |
| VANet [88] | 66.3 | 89.8 | 96.0 | 97.3 | 99.1 | 95.1 | 98.8 | 93.0 | 98.2 | |
| MLPL* [3] | 45.1 | 88.3 | 91.1 | 61.1 | 69.8 | 57.3 | 68.4 | 52.4 | 66.2 | Unsupervised Learning |
| DTDN* [119] | 48.0 | 65.9 | 87.7 | 52.8 | 65.3 | 50.7 | 65.2 | 45.7 | 59.6 | |
| VAPC **[123] | 40.3 | 77.4 | 84.6 | - | - | - | - | - | - | |
| VTGAN + ATTNet [115] | - | - | - | 49.5 | 68.7 | 45.2 | 64.0 | 40.8 | 59.0 | |
| SPGAN + ATTNet [115] | - | - | - | 48.3 | 67.2 | 43.4 | 63.0 | 39.5 | 59.0 | |
| CycleGAN + ATTNet [115] | - | - | - | 42.7 | 60.7 | 38.9 | 57.4 | 35.1 | 53.0 | |

## VII. DISCUSSION AND CHALLENGES

To evaluate deep learning-based vehicle ReID methods according to the proposed taxonomy, Table. V presents the performance comparison of state-of-the-art approaches in vehicle ReID using the Veri-776 and VehicleID datasets. Observably, transformer-based feature learning and metric learning have demonstrated superior performance compared to other categories.

The efficiency of MsKAT [5], a knowledge-based method, derives from its use of a multi-scale knowledge-aware transformer whose high performance can be attributed to the strategic integration of the transformer architecture. Moreover, the effectiveness of unsupervised methods, especially domain adaptation techniques, is significantly reduced due to the inherent divergence between the training and test domains. This adaptation between domains poses a significant challenge in practical vehicle ReID tasks.

In addition, to gauge the dataset's complexity and the model's adaptability to real-world challenges, performance evaluations of some of the state-of-the-art approaches on the VERI-Wild and VERI-Wild 2.0 datasets are presented in Tables. VI and VII. The analysis of Tables. V, VI, and VIII shows the high complexity of the VERI-Wild 2.0 dataset, which leads to a decrease in the model's accuracy. Consequently, this dataset is a suitable benchmark for evaluating vehicle recognition models.

TABLE VI
PERFORMANCE OF STATE-OF-THE-ART APPROACHES IN VEHICLE REID ON VERI-776 AND VEHICLEID DATASETS.

| Models | Small | | | Medium | | | Large | | |
|---|---|---|---|---|---|---|---|---|---|
| | mPA | R1 | R5 | mPA | R1 | R5 | mAP | R1 | R5 |
| OIM [135] | 14.4 | 48.7 | 66.6 | 12.6 | 45.0 | 60.9 | 10.0 | 38.8 | 54.4 |
| BUC [136] | 15.2 | 37.5 | 53.0 | 14.8 | 33.8 | 51.1 | 9.2 | 25.2 | 41.6 |
| SSML [94] | 23.7 | 49.6 | 71.0 | 20.4 | 43.9 | 64.9 | 15.8 | 34.7 | 55.4 |
| VAPC [123] | 33.0 | 72.1 | 87.7 | 28.1 | 64.3 | 82.0 | 22.6 | 55.9 | 75.9 |
| MAPLD [29] | 36.6 | 72.1 | 87.6 | 33.4 | 66.2 | 84.5 | 27.7 | 55.9 | 77.3 |
| DRDL [10] | 22.5 | 57.0 | 75.0 | 19.3 | 51.9 | 71.0 | 14.8 | 44.6 | 61.0 |
| GSTE [95] | 31.4 | 60.5 | 80.1 | 26.2 | 52.1 | 74.9 | 19.5 | 45.4 | 66.5 |
| FDA-Net [128] | 35.1 | 64.0 | 82.8 | 29.8 | 57.8 | 78.3 | 22.8 | 49.4 | 70.5 |
| DFLNet [137] | 66.2 | - | - | 58.2 | - | - | 47.1 | - | - |
| AAVER [49] | 62.2 | 75.8 | 92.7 | 53.6 | 68.2 | 88.8 | 41.6 | 58.6 | 81.5 |
| SAVER [132] | 80.9 | 94.5 | 98.1 | 75.3 | 92.7 | 97.4 | 67.7 | 89.5 | 95.8 |
| PCRNet [22] | 81.2 | 92.5 | - | 75.3 | 89.3 | - | 67.1 | 85.0 | - |
| VARID [97] | 75.4 | 75.3 | 95.2 | 70.8 | 68.8 | 91.8 | 64.2 | 63.2 | 83.2 |
| DFNet [127] | 83.0 | 94.7 | 98.0 | 77.2 | 93.2 | 97.4 | 69.8 | 89.3 | 96.0 |
| SRF [138] | 82.7 | 95.1 | 98.6 | 77.9 | 93.8 | 98.0 | 70.2 | 90.6 | 97.9 |
| GAN-Siamese [118] | 92.5 | - | - | - | - | - | - | - | - |

TABLE VII
PERFORMANCE OF STATE-OF-THE-ART APPROACHES IN VEHICLE REID ON VERI-776 AND VEHICLEID DATASETS.

| Models | Test Set All | | | Test Set A | | | Test Set B | | |
|---|---|---|---|---|---|---|---|---|---|
| | mPA | CMC@1 | CMC@5 | mPA | CMC@1 | CMC@5 | mAP | CMC@1 | CMC@5 |
| GSTE [95] | 32.6 | 59.2 | 64.5 | 33.0 | 47.5 | 50.8 | 41.8 | 86.1 | 91.4 |
| FDA-Net [128] | 34.2 | 57.3 | 64.9 | 34.6 | 45.5 | 52.8 | 43.9 | 84.8 | 92.5 |
| EVER [139] | 36.8 | 59.1 | 67.6 | 36.8 | 48.7 | 57.3 | 45.4 | 86.1 | 94.3 |
| PVEN [133] | 37.2 | 61.2 | 68.6 | 38.8 | 51.3 | 59.3 | 45.5 | 88.0 | 94.4 |
| SAVER [132] | 38.0 | 62.1 | 69.5 | 39.2 | 52.3 | 60.2 | 45.1 | 88.1 | 94.1 |
| DFNet [127] | 39.8 | 62.2 | 68.9 | 40.4 | 51.7 | 60.5 | 46.1 | 88.6 | 94.2 |
| GAN-Siamese [118] | 68.5 | - | - | - | - | - | - | - | - |

The widespread adoption of intelligent video surveillance systems has led to a substantial surge in the demand for vehicle ReID. Despite considerable and sustained efforts over an extended period, this field mainly confronts two significant challenges. First, the intra-instance discrepancy observed in images of the same vehicle captured in different modes, including variations in camera views, vehicle perspectives, and capture times. Second, the inter-instance similarity encountered among different vehicles, notably when sharing the same characteristics such as color, type, and manufacturer. Furthermore, in applying vehicle ReID within traffic surveillance scenes, variations in image resolution, diverse camera angles, weather conditions, and disparate lighting conditions among different cameras lead to significant differences in vehicle appearance, posing substantial challenges for vehicle ReID. The complexities and challenges facing the vehicle ReID problem are explained below.

### A. Viewpoint Variability

Vehicles can appear drastically different based on camera angles, distances, and orientations. Extracting viewpoint invariant features to overcome these viewpoint changes to ensure consistent identification remains challenging.

### B. Appearance Variations

External factors such as lighting conditions, weather, occlusions, or modifications (addition/removal of accessories) can alter a vehicle's appearance, making consistent identification challenging.

### C. Scale and Resolution

Images of vehicles may differ in resolution and scale across cameras, impacting feature extraction and matching accuracy.

### D. Intra-Class Variability

Vehicles of the same make and model can exhibit significant visual differences due to modifications, different versions, or wear and tear, complicating differentiation between similar vehicles.

### E. Limited Annotated Data

Despite the diversity of different data sets, annotated data that covers all possible situations is still insufficient and creates a significant challenge for training robust models and hinders the development of accurate and generalizable algorithms.

### F. Real-time Processing

Obviously, deep learning models face exponential-order time complexities, which is a persistent and significant obstacle for their practical application in real-time applications. As a result, accurate re-identification of vehicles in real-time traffic monitoring or management systems requires the deployment of algorithms that have fast processing capabilities.

### G. Privacy and Ethical Concerns

Balancing the effectiveness of re-identification while respecting privacy rights, especially in public surveillance, presents a critical challenge.

## VIII. CONCLUSION

The field of Vehicle ReID focuses on establishing associations among vehicle images obtained from a distributed network of cameras operating across diverse traffic environments. The increased adoption of intelligent video surveillance systems has precipitated a considerable demand for advancements in Vehicle ReID. This task holds paramount significance within the domain of vehicle-centric technologies, serving as a crucial catalyst in the implementation of ITS and the progression of initiatives aimed at developing smart cities. Recent strides in deep learning have markedly catalyzed the evolution of Vehicle ReID technologies. This paper offers an exhaustive exploration of deep learning methodologies as applied to Vehicle ReID. The methodologies outlined herein are classified into two primary categories: supervised and unsupervised approaches. Within supervised methods, models have predominantly focused on acquiring distinctive visual features from vehicle images, treating it as a classification problem. Conversely, other models have prioritized deep metric learning through the application of specific loss functions. Unsupervised approaches seek to extract pertinent information from data without recourse to class labels, bifurcating into two distinct groups: unsupervised domain adaptation and fully unsupervised methods. This paper undertakes a comprehensive taxonomy review, elucidating these methodologies. Additionally, a comparative analysis of established standard datasets and evaluation criteria is presented, accompanied by a meticulous comparison of experimental results from state-of-the-art papers. This paper aims to serve as a guiding framework and a valuable resource for prospective research endeavors in this domain.

## ACKNOWLEDGMENT

This study is supported by Scientific and Technological Research Council of Türkiye (TUBITAK) with 2221 - Fellowships for Visiting Scientists and Scientists on Sabbatical Leave Support Program; and Hacettepe University Scientific Research Project Department with code FAY-2022-20118.